\documentclass[runningheads]{llncs}

\usepackage[T1]{fontenc}
\usepackage{cite}
\usepackage{amsmath,amssymb,amsfonts}
\usepackage{textcomp}
\usepackage{xcolor}
\usepackage{hyperref}
\usepackage{todonotes}
\usepackage{fancyhdr}
\usepackage{enumitem}
\usepackage{algorithm}
\usepackage{algorithmic}
\usepackage{graphicx}
\usepackage{subcaption}
\usepackage{booktabs}
\usepackage{multirow}
\usepackage{makecell}
\begin{document}

\newif\ifdraft
%\draftfalse
\drafttrue

\ifdraft
\newcommand{\DA}[1]{{\color{red}\textbf{DA: #1}}}
\newcommand{\da}[1]{{\color{red} #1}}
\else
\newcommand{\DA}[1]{}
\newcommand{\da}[1]{}
\fi

% ==== TITLE ======

% \title{A Visual Temporal Policy for Dribbling in Humanoid Robot Soccer}
\title{Vision-Based Dribbling for Humanoid Soccer via Privileged Representation Learning}
\author{
Flavio Maiorana\inst{1}\orcidID{0009-0003-2059-7254} \and
Valerio Spagnoli\inst{1}\orcidID{0009-0008-0284-9602} \and
Eugenio Bugli\inst{1}\orcidID{0009-0000-9540-681X} \and
Flavio Volpi\inst{2}\orcidID{0009-0004-9822-5124} \and
Daniele Affinita\inst{3}\orcidID{0009-0000-9347-9847} \and
Vincenzo Suriani\inst{1}\orcidID{0000-0003-1199-8358} \and
Daniele Nardi\inst{1}\orcidID{0000-0001-6606-200X} \and
Luca Iocchi\inst{1}\orcidID{0000-0001-9057-8946}
}
\institute{
Dept. of Computer, Control, and Management Engineering\\ Sapienza University of Rome, Rome (Italy),
\email{\{lastname\}@diag.uniroma1.it} 
\and
Institut de Robòtica i Informàtica Industrial (CSIC-UPC), C/ Llorens i Artigas 4-6, 08028, Barcelona, Spain,
\email {fvolpi@iri.upc.edu}
\and
École Polytechnique Fédérale de Lausanne (EPFL), Lausanne, Switzerland,
\email{\{lastname\}@epfl.ch}
}

\maketitle

\captionsetup[table]{skip=10pt}
\renewcommand{\arraystretch}{1.3} % default is 1.0

% ===== MAIN BODY ======

\begin{abstract}
Recent advances in humanoid robotics have highlighted the importance of
deployable loco-manipulation skills. Dribbling a soccer ball while
evading active opponents requires simultaneous balance, precise ball control,
and awareness of a dynamic adversary under onboard sensing and real-time
constraints. Existing approaches typically separate perception and motion,
which can be effective in controlled settings but may fail under occlusions,
fast ball movements, and complex opponent interactions, since perception is not
directly optimized for control.

We propose an integrated approach in which a temporal depth encoder is embedded
into a reinforcement learning policy through a task-specific projection layer.
We apply this framework to a simulated Booster T1 humanoid robot and show that it is possible to learn vision-based, opponent-aware dribbling directly from depth
observations, without explicit state estimation or privileged scene
information. The learned policy achieves \(100\%\) success in nominal
target-driven dribbling and \(96\%\) success with a single static obstacle,
while reaching \(46\%\) success against an actively moving ball-attacker
opponent. These results demonstrate that the proposed framework supports robust
vision-based dribbling in nominal and moderately dynamic settings, and provides
a strong foundation for handling more challenging moving-adversary scenarios.

\end{abstract}

\section{Introduction}
\label{sec:introduction}

Reinforcement learning has enabled legged robots to master complex locomotion behaviors across challenging terrains \cite{kumar_rma_2021, zhuang_robot_2023}, and more recently to manipulate objects in the environment while walking - a setting that demands simultaneous control of balance, contact forces, and object dynamics \cite{fu_deep_2022}.
However, despite the progress that DRL has allowed, loco-manipulation task still remains an open challenge for both quadruped and bipedal robots. If the problem results complex for the first, and a wide range of solutions have been explored \cite{liu_visual_2024, yokoyama_asc_2023}, for the second the challenge is still harder, due to the mixture of coordination, balance and stability that are needed to execute a task \cite{gu_humanoid_2026}.  

Soccer dribbling is a particularly demanding instance of this problem: the robot must maintain continuous ball control while navigating toward a goal, adapting its gait to ball velocity, and reacting to perturbations. Previous works have demonstrated dribbling on quadrupeds \cite{ji_dribblebot_2023, hu_dexdribbler_2024} on diverse terrains, but these systems either rely on human-provided velocity commands or assume that the robot acts in isolation, without an adversary. Compared with quadrupeds, humanoids must also preserve dynamic balance under the perturbations introduced by ball contact.

The soccer setting, especially considering the existence of rapid ball movements and active opponents, fundamentally changes the perception requirements. The robot must now track not only the ball but also the opponent's position relative to itself, and it must maintain this estimate even when the ball is momentarily occluded - a frequent occurrence on a humanoid, where the ball regularly disappears during close-range dribbling. This challenge has grown qualitatively harder as humanoid hardware has improved. 
Modern RoboCup humanoid platforms, such as the Booster T1, operate at higher speeds than earlier platforms, increasing the perception and control demands.

The dominant paradigm for addressing the ball control in humanoid soccer has been to decouple perception from control: a detection module (e.g. YOLO-based) localizes the ball in each frame, and a Kalman filter propagates the state estimate across frames and through missed detections \cite{hu_humanoid_2023}. While effective in controlled settings, this modular design has a fundamental limitation for learning-based control: the perception module is optimized independently of the policy, with no guarantee that its output representation is sufficient for the downstream task. 

In this work, we address these challenges by learning perception and control jointly through a temporal encoder whose representation is shaped by what the policy actually needs to act well.
We formulate the task as an obstacle-aware dribbling problem and address it with a perception-integrated RL system inspired by Rapid Motor Adaptation (RMA) \cite{kumar_rma_2021}: a privileged encoder provides rich state information during training, which a deployable temporal encoder learns to approximate at inference time. Integrated with this representation, we train a policy under a curriculum that progressively increases opponent difficulty and with a reward shaping system that encourages ball possession and opponent avoidance. All experiments are conducted in simulation on the Booster T1 humanoid platform using mjlab \cite{zakka_mjlab_2026}, and we outline a principled sim-to-real roadmap as future work.
The contributions of this paper are threefold:
\begin{itemize}
    \item We propose a temporal encoder trained through privileged distillation to infer task-relevant ball and opponent states from onboard depth observations, enabling deployable perception without explicit state estimation.
    \item We introduce an obstacle-aware training pipeline that progressively exposes the robot to increasingly challenging dribbling scenarios, from nominal ball control to static blockers and dynamic opponent-like obstacles.
    \item We present and release an end-to-end reinforcement learning system for autonomous humanoid soccer dribbling, integrating perception, balance control, ball handling, and opponent avoidance within a unified policy framework. Code and extra-material are released at \url{https://lab-rococo-sapienza.github.io/learning-to-dribble/}.
\end{itemize}
The remainder of the paper is organized as follows. Section \ref{sec:rel_work} reviews related work, Section \ref{sec:methodology} presents the proposed methodology, Section \ref{sec:results} reports the experimental results, and Section \ref{sec:conclusion} concludes the paper.
\section{Related Work}
\label{sec:rel_work}
Recent advances in deep reinforcement learning (RL) have enabled legged robots to acquire robust and adaptive locomotion skills across a wide range of terrains. \cite{kumar_rma_2021} introduced an adaptation module that infers latent environment parameters online, allowing for walking with terrain variation and payload changes, and \cite{kumar_adapting_2022} extended the approach to bipedal robots. These works highlight the importance of latent adaptation for handling partial observability and dynamics uncertainty in locomotion. \cite{margolis_rapid_2022} pushes this further with an end-to-end controller characterized by an adaptive curriculum on velocity commands and online system identification, while \cite{li_learning_2023} combines experience replay with an automatic curriculum strategy to achieve agile and adaptive behaviors including fall recovery and high-speed turning. Along a complementary direction, \cite{wang_cts_2024} proposes a concurrent teacher-student scheme that jointly trains privileged and deployable policies, improving sample efficiency and sim-to-real robustness in challenging environments. On the bipedal side, \cite{li_reinforcement_2024} introduces a unified RL framework with a dual-history architecture that enables the biped to walk, run, and jump, identifying task randomization as a key source of robustness.
Several works have targeted agile locomotion over challenging terrains such as \cite{hoeller_anymal_2023} \cite{rudin_parkour_2025}. \cite{chen_vmts_2025} integrates visual information into a teacher-student approach for robust bipedal locomotion over uneven terrains, using a mixture-of-experts policy with an alignment loss between teacher and student networks.

% \paragraph{Learning-Based Loco-Manipulation and Robot Soccer}
Soccer and dribbling tasks represent a challenging class of loco-manipulation problems, requiring simultaneous control of locomotion and object dynamics.  Early systems relied on modular, model-based pipelines with limited adaptability \cite{bianchi_ball_2015}.
On quadrupeds, DribbleBot \cite{ji_dribblebot_2023} learns dynamic ball manipulation directly from sensory observations, while DexDribbler \cite{hu_dexdribbler_2024} achieves dexterous ball control through RL with dynamic supervision. On the other hand, hierarchical frameworks decouple kicking from target-aware planning for precise shooting, leveraging deep reinforcement learning to train both robust motion control policies and planning policies that enable accurate real-world ball targeting \cite{ji_hierarchical_2022}.
Humanoid soccer introduces additional challenges due to dynamic balance, higher degrees of freedom, and increased sensitivity to perturbations. \cite{haarnoja_learning_2024} trains a humanoid to play soccer end-to-end, with agile kicking, fall recovery, and tactical behaviors emerging from self-play, while \cite{tirumala_learning_2024} removes reliance on privileged state by training soccer policies directly from RGB vision. More recently, \cite{wang_learning_2025} presents a unified RL controller for humanoid soccer that tightly integrates visual perception with motion control. Complementary work such as \cite{chen_hifar_2025} addresses robust fall recovery through multi-stage curriculum learning.

In contrast, our work investigates humanoid dribbling in a dynamic environment, where obstacles mimicking opponent behaviors force the policy to plan around interferences, rather than operate in a free field. We build on the RMA paradigm with a temporal-depth encoder that jointly distills perception and control, producing a deployable visual adaptation module trained end-to-end that addresses both the humanoid loco-manipulation gap and the lack of adversaries in prior benchmarks.

\section{Methodology}
\label{sec:methodology}

\noindent \textbf{Problem Formulation.} 
We consider a dribbling task where a humanoid robot must maintain ball control while avoiding opponents. 
The problem is formulated as a Partially Observable Markov Decision Process (POMDP).
$$
\mathcal{M}=(\mathcal{S},\mathcal{O},\mathcal{A},p,r,\gamma)
$$

The state space $\mathcal{S}$ includes the full world-frame state of all entities, namely the robot, ball, and opponent obstacles, as well as internal command variables such as the target position. The observation space $\mathcal{O}$ comprises both proprioceptive signals (e.g., IMU readings and joint positions) and exteroceptive information (e.g., ball position, ball velocity, and obstacle positions). In particular, the exteroceptive information is encoded into a latent vector $z_t \in \mathbb{R}^{d_z}$ that summarizes historical observations \cite{kumar_rma_2021}.
The latent vector $z_t$ is produced according to the two-phase training scheme detailed in Sec.~\ref{sec:VEL}. 

The action space $\mathcal{A}$ contains a joint position vector $a_t \in \mathbb{R}^{21}$. The vector represents only the body joints, we assume the head is controlled independently. The transition function $p(s_{t+1} \mid s_t, a_t)$ is induced by the physics simulator \footnote{\cite{zakka_mjlab_2026}} and task randomization, and is assumed to be unknown to the policy. As for the rewards, the task-related terms include (i) ball-velocity tracking, (ii) robot-ball proximity, (iii) robot-ball directional alignment, (iv) target progress, and (v) target completion, while 
additional penalties discourage unsafe obstacle interactions, while regularization
terms promote feasible locomotion, smooth actions, stable contacts, and consistent
gait patterns.

The policy receives only a partial observation $o_t \in \mathcal{O}$. Following the RMA formulation, the actor observation is composed of proprioceptive terms together with a latent vector representing exteroceptive information. Formally, the actor observation is:
\begin{equation}
    o_{t}^{\text{actor}} = \left[ o_t^{\mathrm{prop}}, z_t \right]
\end{equation}
where the proprioceptive component $o_t^{\mathrm{prop}}$ is composed by:
\begin{itemize}
    \item base angular velocity from the IMU, $\boldsymbol{\omega}_t^{base}$;
    \item projected gravity in the robot body frame, $\mathbf{g}_t^{proj}$;
    \item relative joint positions, $\mathbf{q}_t^{rel}$;
    \item relative joint velocities, $\dot{\mathbf{q}}_t^{rel}$;
    \item previous action, $\mathbf{a}_{t-1}$;
    \item gait-phase command, $\boldsymbol{\phi}_t$;
    \item commanded ball velocity expressed in the robot body frame, $\mathbf{v}_t^{ball}$;
\end{itemize}

The proprioceptive component includes IMU measurements, projected gravity,
relative joint positions and velocities, the previous action, gait phase, and the
commanded ball velocity expressed in the robot frame.

Instead, the latent variable $z_t\in \mathbb{R}^{d_z}$ provides additional exteroceptive information derived from high-dimensional sensory inputs (e.g., raw obstacle or ball detections), ensuring that the actor never directly observes privileged simulator states.

\noindent \textbf{Curriculum Learning.} 
Training progresses along two axes: (i) \emph{phase} defining how the latent exteroceptive representation $z_t$ is obtained, (ii) \emph{stage}, defining obstacle configuration difficulty.

The task is specified by a persistent target in the world frame $\mathbf{g}^w \in \mathbb{R}^2$ sampled around the ball position $\mathbf{b}^w_t$, avoiding velocity resampling at every step.

\begin{figure}[t]
    \centering
    \includegraphics[width=\textwidth]{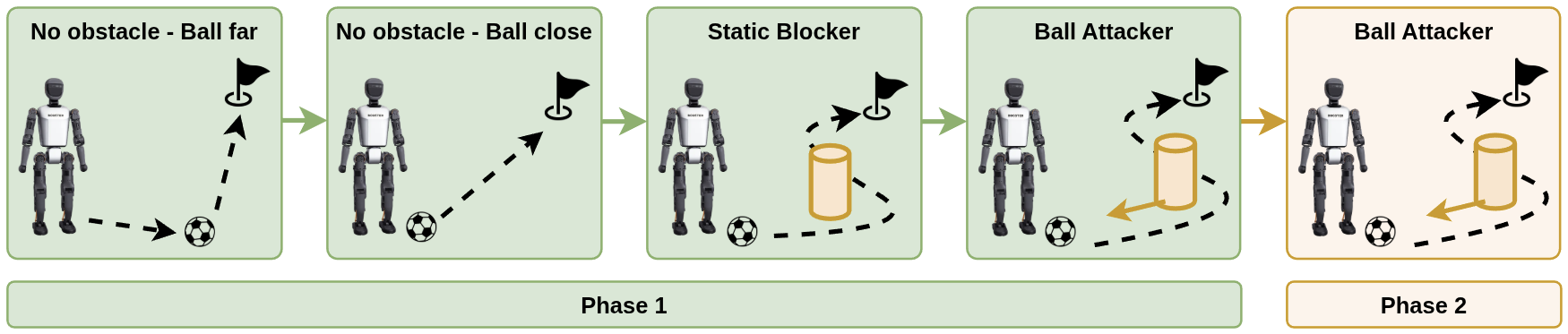}
    \caption{\textbf{Curriculum-based training schedule.} The policy is first trained through progressively harder stages, from ball approach to dynamic opponent interaction, and then a visual encoder is trained to reproduce the learned latent from depth observations, while keeping the policy frozen.}
    \label{fig:curriculum}
\end{figure}

\paragraph{Phase 1 (policy learning).} A four-stage curriculum increases dribbling difficulty as depicted in Figure \ref{fig:curriculum}.
\emph{Stage 0:} no obstacles, ball far from the robot. The agent learns approach and dribbling.
\emph{Stage 1:} no obstacles, ball near the robot. The agent refines target-directed dribbling.
\emph{Stage 2:} one static blocker on the ball-target corridor. The agent acquires obstacle avoidance and trajectory recovery. 
\emph{Stage 3:} one dynamic opponent on the ball-target corridor interacting with the ball. The agent learns online reactive control.

Different optimization strategies are used for different stages. In particular, Stage 0 uses plain PPO, while stages 1-3 are initialized from the respective previous stage checkpoint and use a DAgger-regularized PPO objective \cite{ross_reduction_2010} with the stage-0 policy $\pi_0$ as a teacher: 

\[
\mathcal{L}_{\mathrm{total}}
=
\mathcal{L}_{\mathrm{PPO}}
+
\lambda_{\mathrm{imit}} \mathbb{E}_{s_t}\left[ \left|\left| \pi_\theta(a_t \mid s_t) - \pi_0(a_t \mid s_t) \right|\right|_2^2 \right].
\]

The imitation component mitigates catastrophic forgetting, preserving the nominal locomotion and ball-handling strategy learned in Stage 0.

\paragraph{Phase 2 (visual adaptation)}
Initialized from the Phase 1 Stage 3 checkpoint, the policy is frozen, and a visual encoder is trained to reconstruct the latent $z_t$ from depth observations.

\begin{figure}[t]
    \centering
    \includegraphics[width=\textwidth]{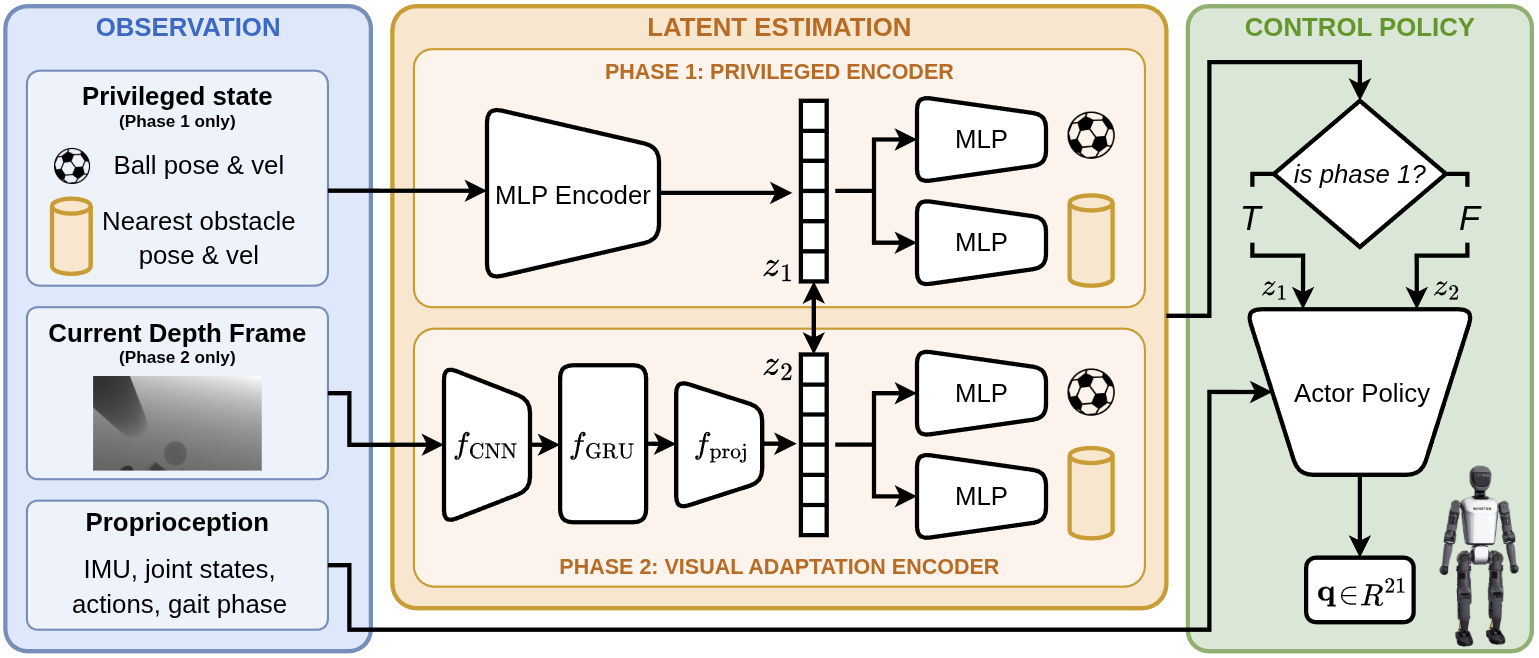}
    \caption{\textbf{Overview of the two-phase framework}. In Phase 1, a privileged encoder maps ground-truth ball and obstacle states to a latent representation used by the policy. In Phase 2, a visual encoder predicts the same latent from depth observations, enabling deployment with onboard sensing only.}
    \label{fig:architecture}
\end{figure}

\noindent \textbf{Visual Encoder Learning.} 
\label{sec:VEL} 
To address partial observability, Phase~2 replaces the privileged latent generator with a visual adaptation module that predicts the same latent from depth observations. The actor input remains unchanged: proprioception concatenated with a latent $z_t \in \mathbb{R}^{d_z}$ encoding task-relevant exteroceptive information. The two phases differ only in how $z_t$ is produced (Fig.~\ref{fig:architecture}).

\paragraph{Phase 1: Privileged encoder.}
In the first phase, the latent is generated by a privileged encoder with access to ground-truth task variables, including the state of the ball and the state of the nearest dynamic obstacle. These quantities are represented in the robot coordinate frame and mapped by a small MLP to a latent vector $z_t$, which is concatenated with the proprioceptive observation and provided to the policy.

\paragraph{Phase 2: Visual adaptation encoder.}
In the second phase, the privileged encoder is replaced by a visual adaptation module that aims to predict the same latent vector from onboard depth sensing only.
Specifically, we use a convolutional neural network (CNN) encoder followed by a gated recurrent unit (GRU). The convolutional layers extract spatial features from each depth frame, while the GRU aggregates them over time to capture temporal continuity and short-horizon motion cues. The recurrent output is then projected into the same latent space used in Phase 1.

Formally, given the current depth frame $\mathbf{D}_t$ resized to $108 \times 192$ preserving the original aspect ratio, the visual adaptation module computes
\[
\mathbf{e}_t = f_{\mathrm{cnn}}(\mathbf{D}_t) \in \mathbb{R}^{64}, \qquad
\mathbf{h}_t = f_{\mathrm{gru}}(\mathbf{e}_t,\mathbf{h}_{t-1}) \in \mathbb{R}^{256}, \qquad
z_t = f_{\mathrm{proj}}(\mathbf{h}_t) \in \mathbb{R}^{64},
\]
where $f_{\mathrm{cnn}}$ is the convolutional encoder, $f_{\mathrm{gru}}$ is the recurrent temporal aggregator, and $f_{\mathrm{proj}}$ maps the recurrent state to the policy latent space.

The visual encoder is trained to match the privileged latent through the loss
\[
\mathcal{L}_{\mathrm{latent}} = \|z_t^{\mathrm{adapt}} - z_t^{\mathrm{priv}}\|_2^2.
\]
To improve identifiability and stabilize learning, auxiliary prediction heads also regress the ball state and nearest-obstacle state from the latent:
\[
\mathcal{L}_{\mathrm{adapt}}
=
\mathcal{L}_{\mathrm{latent}}
+
\lambda_{\mathrm{ball}}^{\mathrm{pos}}\mathcal{L}_{\mathrm{ball}}^{\mathrm{pos}}
+
\lambda_{\mathrm{ball}}^{\mathrm{vel}}\mathcal{L}_{\mathrm{ball}}^{\mathrm{vel}}
+
\lambda_{\mathrm{obs}}^{\mathrm{pos}}\mathcal{L}_{\mathrm{obs}}^{\mathrm{pos}}
+
\lambda_{\mathrm{obs}}^{\mathrm{vel}}\mathcal{L}_{\mathrm{obs}}^{\mathrm{vel}}.
\]

This design preserves a stable actor interface across both phases: only the
mechanism used to produce the latent changes, while the policy architecture
itself remains unchanged.

\section{Experimental Results}
\label{sec:results}

\subsection{Simulation Setup and Evaluation Protocol}

All experiments are conducted in simulation using the \texttt{mjlab} framework, on the Booster T1 humanoid dribbling a free soccer ball on a flat terrain.

The policy receives proprioceptive measurements concatenated with a latent vector $z_t$ produced by the visual adaptation module. At deployment, $z_t$ is inferred from a head-mounted depth camera.

We evaluate the dribbling policy under three conditions:
\begin{enumerate}
\item \textbf{No obstacles}: target-reaching dribbling without adversaries.
\item \textbf{Static obstacle}: a single stationary obstacle placed near the
ball-to-target corridor.
\item \textbf{Ball attacker}: a single moving obstacle spawned near the
ball-to-target corridor and scripted to chase the ball at a fixed speed.
\end{enumerate}

At each trial, a target is generated in the horizontal plane conditioned on the initial robot-ball configuration. Target placement, obstacle placement, and (for the dynamic case) obstacle speed are sampled independently within predefined ranges; the resulting scene is fixed for the duration of each trial. Obstacle speed is drawn from $[0.1, 0.4]\,\mathrm{m/s}$.

A trial terminates only upon successful target reach (ball-target
distance \(\le 0.75\,\mathrm{m}\)), timeout (\(30\,\mathrm{s}\)), robot
fall, or ball lost (robot-ball distance above \(2.0\,\mathrm{m}\)). Obstacle
collisions do not terminate the trial and are instead recorded as safety
metrics. The weights associated with each component of the reward function are outlined in Table \ref{tab:reward_weights}

\begin{table}[t]
\centering
\renewcommand{\arraystretch}{1.15}
\begin{tabular}{lc @{\hspace{0.8cm}} lc}
\toprule
\textbf{Reward Term} & \textbf{Weight} & \textbf{Reward Term} & \textbf{Weight} \\
\midrule
Ball velocity tracking & 3.0 & Ball-target reached & 50.0 \\
Ball speed tracking & 2.0 & Robot-obstacle collision & $-10.0$ \\
Ball heading tracking & 2.0 & Ball-obstacle collision & $-10.0$ \\
Robot-ball distance & 0.05 & Foot slip & $-0.1$ \\
Robot-ball yaw & 2.0 & Upright & 1.0 \\
Ball-target progress & 2.0 & Body angular velocity & $-0.05$ \\
Swing phase & 3.0 & Angular momentum & $-0.5$ \\
Stance phase & 2.0 & DoF position limits & $-1.0$ \\
Pose arms & 1.0 & Action rate (L2) & $-0.1$ \\
Pose legs & 1.0 & Foot swing height & $-0.25$ \\
Feet distance & $-6.0$ & Soft landing & $-1 \times 10^{-5}$ \\
Foot-foot contact & $-5.0$ & Self-collisions & $-1.0$ \\
Nonfoot-ball contact & $-2.0$ &  &  \\
\bottomrule
\end{tabular}
\caption{Reward terms and weights for training the dribbling policy.}
\label{tab:reward_weights}
\end{table}

We report the success rate (SR), time-to-target over successful
trials in seconds (T2T), timeout-censored time-to-target in seconds (T2T-C),
fall rate (FR), ball-lost rate (LR), robot-obstacle collision rate (RCR),
robot-obstacle contacts per trial (RC/t), ball-obstacle collision rate (BCR),
ball-obstacle contacts per trial (BC/t), and the minimum ball-obstacle clearance in
meters (MBC). The censored time-to-target assigns failed trials the full
timeout horizon, preventing low-success policies from appearing artificially
fast.

We also report a ball-velocity diagnostic on the same episodes. Let \(v_{\mathrm{ball}}\) denote the realized ball velocity and \(v_{\mathrm{cmd}}\) the commanded ball velocity. We define the vector tracking error as \(e_{\mathrm{vec}} = \lVert v_{\mathrm{ball}} - v_{\mathrm{cmd}} \rVert\), the speed error as \(e_{\mathrm{spd}} = \left| \lVert v_{\mathrm{ball}} \rVert - \lVert v_{\mathrm{cmd}} \rVert \right|\), and the angular error as \(e_{\mathrm{ang}} = \angle\!\left(v_{\mathrm{ball}},\, v_{\mathrm{cmd}}\right)\). In obstacle settings, a good policy may intentionally deviate from the commanded ball velocity to avoid the obstacle while still completing the task successfully.

To distinguish nominal tracking from deliberate avoidance, timesteps in the
obstacle conditions are split into \emph{blocked} and \emph{unblocked}
subsets. A timestep is marked as blocked if at least one active obstacle lies
ahead of the ball along the ball-to-target direction, remains before the
target, is within \(2.0\,\mathrm{m}\) in forward distance, and lies within
\(0.75\,\mathrm{m}\) of the ball-to-target line. Otherwise, it is marked as
unblocked. 

Finally, we report perception metrics on the same episodes to assess the visual
adaptation module during task execution: ball position error \(e_{\mathrm{ball,pos}}\), ball velocity error
\(e_{\mathrm{ball,vel}}\), obstacle position error \(e_{\mathrm{obs,pos}}\),
obstacle velocity error \(e_{\mathrm{obs,vel}}\), and field-of-view coverage
\(c_{\mathrm{fov}}\).

Evaluation is run over 5 random seeds. For each seed, we collect 10 trials per
condition and aggregate results across seeds, for a total of 50 trials per
condition.

\subsection{Dribbling Policy Evaluation}

We first evaluate the final deployable policy, i.e., the policy trained with the full pipeline and executed with the depth-based adaptation encoder. Table \ref{tab:main_eval} summarizes the main task metrics across the three evaluation
conditions.

Table \ref{tab:velocity_eval} reports the velocity diagnostic. In the obstacle-free condition, all timesteps are effectively unblocked. In the two obstacle conditions, we additionally separate the metrics into unblocked and
blocked subsets in order to distinguish nominal tracking from obstacle-avoidance behavior.

Table \ref{tab:perception_eval} reports the perception metrics collected during evaluation. These metrics are computed only on timesteps in which the ball lies within the camera field of view. When the ball is outside the field of view, the visual adaptation module is not used to estimate the ball state, and ground-truth ball position is provided instead.

\begin{table}[t]
\centering
\begin{tabular}{lcccccccccc}
\hline
Condition & SR & T2T & T2T-C & FR & LR & RCR & RC/t & BCR & BC/t & MBC \\[-0.35em]
& [\%] & [s] & [s] & [\%] & [\%] & [\%] & [\#/trial] & [\%] & [\#/trial] & [m] \\
\hline
No obstacles & 100.00 & 11.45 & 11.45 & 0.00 & 0.00 & -- & -- & -- & -- & -- \\
Static obstacle & 96.00 & 13.29 & 13.95 & 4.00 & 0.00 & 8.00 & 0.08 & 4.00 & 0.10 & 1.73 \\
Ball attacker & 46.00 & 11.82 & 21.64 & 52.00 & 2.00 & 68.00 & 0.72 & 40.00 & 0.70 & 1.36 \\
\hline
\end{tabular}
\caption{Final-policy main task evaluation.}
\label{tab:main_eval}
\end{table}

\begin{table}[t]
\centering
\begin{tabular}{llccc}
\hline
% Condition & Segment & Vector error & Speed error & Angular error \\ [-0.35em]
Condition & Segment & $e_{\mathrm{vec}}$ [m/s] & $e_{\mathrm{spd}}$ [m/s] & $e_{\mathrm{ang}}$ [deg] \\
\hline
No obstacles & All timesteps & 0.71 & 0.62 & 28.39 \\
Static obstacle & Unblocked & 0.74 & 0.62 & 35.62 \\
Static obstacle & Blocked & 0.74 & 0.60 & 33.96 \\
Ball attacker & Unblocked & 0.74 & 0.61 & 34.27 \\
Ball attacker & Blocked & 0.79 & 0.60 & 43.00 \\
\hline
\end{tabular}
\caption{Velocity-tracking diagnostic for the final policy.}
\label{tab:velocity_eval}
\end{table}

\begin{table}[t]
\centering

\begin{minipage}[t]{0.56\linewidth}
\centering
\begin{tabular}{lccccc}
\hline
Condition & $e_{\mathrm{ball,pos}}$ & $e_{\mathrm{ball,vel}}$ & $e_{\mathrm{obs,pos}}$ & $e_{\mathrm{obs,vel}}$ & $c_{\mathrm{fov}}$ \\[-0.35em]
 & [m] & [m/s] & [m] & [m/s] & [\%] \\
\hline
\makecell[l]{No\\obstacles} & 0.05 & 0.25 & -- & -- & 75.43 \\
\makecell[l]{Static\\obstacle} & 0.05 & 0.24 & 1.26 & 0.21 & 76.74 \\
\makecell[l]{Ball\\attacker} & 0.08 & 0.26 & 1.26 & 0.13 & 78.39 \\
\hline
\end{tabular}
\caption{Perception metrics for the final policy.}
\label{tab:perception_eval}
\end{minipage}
\hspace{3pt} % control horizontal spacing
\begin{minipage}[t]{0.40\linewidth}
\centering
\begin{tabular}{lcccc}
\hline
Condition & Stage & Stage & Stage \\ [-0.35em]
 & 1 & 2 & 3 \\
\hline
\makecell[l]{No\\obstacles} & 100 & 68 & 90 \\
\makecell[l]{Static\\obstacle} & 24 & 42 & 88 \\
\makecell[l]{Ball\\attacker} & 2 & 2 & 46 \\
\hline
\end{tabular}
\caption{Success-rate ablation across curriculum stages in percentage}
\label{tab:curriculum_success}
\end{minipage}

\end{table}

The final policy achieves perfect performance in the no-obstacle condition,
with \(100\%\) success and no failures, showing that nominal target-driven
dribbling is reliably preserved. Performance remains strong in the static
obstacle setting, where success stays high at \(96\%\) with only a modest
increase in time-to-target and low collision rates, indicating effective
handling of a single stationary obstacle. In contrast, the ball-attacker
condition remains substantially more challenging: success drops to \(46\%\),
while both fall and collision rates increase markedly, showing that robust
dribbling against an actively moving adversary is still an open problem.

The velocity diagnostic is consistent with this trend. Tracking errors remain
similar between the no-obstacle and static-obstacle conditions, suggesting that
the policy can avoid a stationary obstacle without strongly disrupting the
commanded ball motion. In the ball-attacker setting, however, the blocked
timesteps exhibit the largest angular and vector errors, indicating stronger
deviations from the nominal command during active avoidance. Perception metrics
remain relatively stable across conditions, with only a small degradation in
ball-state estimation and comparable obstacle-state errors in the two obstacle
settings. This suggests that the main limitation in the hardest condition is
not a severe perception failure, but the difficulty of closed-loop control
against a moving opponent.

\subsection{Ablation: Curriculum Stages}

To study the effect of curriculum training, we evaluate the checkpoint produced after each curriculum stage on all evaluation conditions. As described in Section~\ref{sec:methodology}, the full curriculum consists of four stages. However, in this ablation, we exclude Stage 0, since it only covers a basic prerequisite skill that is shared by all subsequent stages. For each remaining checkpoint, we report the success rate on the three evaluation conditions defined by the current protocol: no obstacles, static obstacle, and ball attacker. Table \ref{tab:curriculum_success}, where each stage has been evaluated over 5 seeds, each with 10 trials, shows the effectiveness of the curriculum strategy, since stage 1 performs poorly when obstacles are present in the scene. 

\section{Conclusion}
\label{sec:conclusion}

This paper presented a visual temporal policy for autonomous humanoid soccer
dribbling. We formulated the task as an opponent-aware loco-manipulation
problem in which the robot must jointly maintain balance, control the ball, and
avoid obstacles.

The experimental results show that the proposed policy reliably solves nominal
target-driven dribbling, achieving \(100\%\) success in the no-obstacle
condition, and also performs well in the single static-obstacle setting, where
success remains high at \(96\%\). Performance degrades substantially in the
ball-attacker setting, where success drops to \(46\%\) and both fall and
collision rates increase markedly. This indicates that the main remaining
challenge is robust closed-loop control against an actively moving adversary,
rather than nominal dribbling itself. The velocity and perception diagnostics
support this interpretation: deviations from the commanded ball motion increase
during blocked attacker interactions, while perception errors remain relatively
stable across conditions.

Overall, these results suggest that combining temporal visual adaptation with
reinforcement learning is a promising approach for humanoid loco-manipulation tasks. However,
the current evidence is limited by the small number of independently trained
policies and by the fact that all experiments are conducted in simulation.

Beyond humanoid soccer, the same framework could support other embodied robotic
tasks requiring perception-driven control under partial observability. Future
work will focus on harder and more diverse opponent behaviors, stronger
robustness against moving adversaries, and sim-to-real transfer on the Booster
T1 humanoid platform.

% ==== Bibliography ====

\bibliographystyle{abbrv}
\bibliography{references}

@article{kumar_adapting_2022,
	title = {Adapting {Rapid} {Motor} {Adaptation} for {Bipedal} {Robots}},
	copyright = {https://doi.org/10.15223/policy-029},
	url = {https://ieeexplore.ieee.org/document/9981091/},
	doi = {10.1109/IROS47612.2022.9981091},
	urldate = {2026-04-17},
	journal = {2022 IEEE/RSJ International Conference on Intelligent Robots and Systems (IROS)},
	author = {Kumar, Ashish and Li, Zhongyu and Zeng, Jun and Pathak, Deepak and Sreenath, Koushil and Malik, Jitendra},
	month = oct,
	year = {2022},
	note = {Conference Name: 2022 IEEE/RSJ International Conference on Intelligent Robots and Systems (IROS)
ISBN: 9781665479271
Place: Kyoto, Japan
Publisher: IEEE},
	pages = {1161--1168},
}

@inproceedings{zhuang_robot_2023,
	title = {Robot {Parkour} {Learning}},
	url = {https://proceedings.mlr.press/v229/zhuang23a.html},
	language = {en},
	urldate = {2026-04-16},
	booktitle = {Proceedings of {The} 7th {Conference} on {Robot} {Learning}},
	publisher = {PMLR},
	author = {Zhuang, Ziwen and Fu, Zipeng and Wang, Jianren and Atkeson, Christopher G. and Schwertfeger, Sören and Finn, Chelsea and Zhao, Hang},
	month = dec,
	year = {2023},
	note = {ISSN: 2640-3498},
	pages = {73--92},
}

@article{kumar_rma_2021,
	title = {{RMA}: {Rapid} {Motor} {Adaptation} for {Legged} {Robots}},
	shorttitle = {{RMA}},
	url = {http://www.roboticsproceedings.org/rss17/p011.pdf},
	doi = {10.15607/RSS.2021.XVII.011},
	urldate = {2026-04-15},
	journal = {Robotics: Science and Systems XVII},
	author = {Kumar, Ashish and Fu, Zipeng and Pathak, Deepak and Malik, Jitendra},
	month = jul,
	year = {2021},
	note = {Conference Name: Robotics: Science and Systems 2021
ISBN: 9780992374778
Publisher: Robotics: Science and Systems Foundation},
}

@misc{zakka_mjlab_2026,
	title = {mjlab: {A} {Lightweight} {Framework} for {GPU}-{Accelerated} {Robot} {Learning}},
	shorttitle = {mjlab},
	url = {https://ui.adsabs.harvard.edu/abs/2026arXiv260122074Z},
	doi = {10.48550/arXiv.2601.22074},
	urldate = {2026-02-23},
	publisher = {arXiv},
	author = {Zakka, Kevin and Liao, Qiayuan and Yi, Brent and Le Lay, Louis and Sreenath, Koushil and Abbeel, Pieter},
	month = jan,
	year = {2026},
	note = {ADS Bibcode: 2026arXiv260122074Z},
}

@article{ji_dribblebot_2023,
	title = {{DribbleBot}: {Dynamic} {Legged} {Manipulation} in the {Wild}},
	copyright = {https://doi.org/10.15223/policy-029},
	shorttitle = {{DribbleBot}},
	url = {https://ieeexplore.ieee.org/document/10160325/},
	doi = {10.1109/ICRA48891.2023.10160325},
	urldate = {2026-04-11},
	journal = {2023 IEEE International Conference on Robotics and Automation (ICRA)},
	author = {Ji, Yandong and Margolis, Gabriel B. and Agrawal, Pulkit},
	month = may,
	year = {2023},
	note = {Place: London, United Kingdom
Publisher: IEEE},
	pages = {5155--5162},
}

@article{hu_dexdribbler_2024,
	title = {{DexDribbler}: {Learning} {Dexterous} {Soccer} {Manipulation} via {Dynamic} {Supervision}},
	copyright = {https://doi.org/10.15223/policy-029},
	shorttitle = {{DexDribbler}},
	url = {https://ieeexplore.ieee.org/document/10802022/},
	doi = {10.1109/IROS58592.2024.10802022},
	urldate = {2026-04-11},
	journal = {2024 IEEE/RSJ International Conference on Intelligent Robots and Systems (IROS)},
	author = {Hu, Yutong and Wen, Kehan and Yu, Fisher},
	month = oct,
	year = {2024},
	note = {Place: Abu Dhabi, United Arab Emirates
Publisher: IEEE},
	pages = {12910--12917},
}

@article{gu_humanoid_2026,
	title = {Humanoid {Locomotion} and {Manipulation}: {Current} {Progress} and {Challenges} in {Control}, {Planning}, and {Learning}},
	volume = {31},
	issn = {1941-014X},
	shorttitle = {Humanoid {Locomotion} and {Manipulation}},
	url = {https://ieeexplore.ieee.org/abstract/document/11456437},
	doi = {10.1109/TMECH.2025.3579247},
	number = {2},
	urldate = {2026-04-25},
	journal = {IEEE/ASME Transactions on Mechatronics},
	author = {Gu, Zhaoyuan and Li, Junheng and Shen, Wenlan and Yu, Wenhao and Xie, Zhaoming and McCrory, Stephen and Cheng, Xianyi and Shamsah, Abdulaziz and Griffin, Robert and Liu, C. Karen and Kheddar, Abderrahmane and Peng, Xue Bin and Zhu, Yuke and Shi, Guanya and Nguyen, Quan and Cheng, Gordon and Gao, Huijun and Zhao, Ye},
	month = apr,
	year = {2026},
	pages = {2300--2330},
}

@misc{yokoyama_asc_2023,
	title = {{ASC}: {Adaptive} {Skill} {Coordination} for {Robotic} {Mobile} {Manipulation}},
	shorttitle = {{ASC}},
	url = {http://arxiv.org/abs/2304.00410},
	doi = {10.48550/arXiv.2304.00410},
	language = {en},
	urldate = {2026-04-25},
	publisher = {arXiv},
	author = {Yokoyama, Naoki and Clegg, Alex and Truong, Joanne and Undersander, Eric and Yang, Tsung-Yen and Arnaud, Sergio and Ha, Sehoon and Batra, Dhruv and Rai, Akshara},
	month = nov,
	year = {2023},
	note = {arXiv:2304.00410 [cs]},
}

@misc{liu_visual_2024,
	title = {Visual {Whole}-{Body} {Control} for {Legged} {Loco}-{Manipulation}},
	url = {http://arxiv.org/abs/2403.16967},
	doi = {10.48550/arXiv.2403.16967},
	language = {en},
	urldate = {2026-04-25},
	publisher = {arXiv},
	author = {Liu, Minghuan and Chen, Zixuan and Cheng, Xuxin and Ji, Yandong and Qiu, Ri-Zhao and Yang, Ruihan and Wang, Xiaolong},
	month = nov,
	year = {2024},
	note = {arXiv:2403.16967 [cs]},
}

@misc{fu_deep_2022,
	title = {Deep {Whole}-{Body} {Control}: {Learning} a {Unified} {Policy} for {Manipulation} and {Locomotion}},
	shorttitle = {Deep {Whole}-{Body} {Control}},
	url = {http://arxiv.org/abs/2210.10044},
	doi = {10.48550/arXiv.2210.10044},
	language = {en},
	urldate = {2026-04-25},
	publisher = {arXiv},
	author = {Fu, Zipeng and Cheng, Xuxin and Pathak, Deepak},
	month = oct,
	year = {2022},
	note = {arXiv:2210.10044 [cs]},
}

@inproceedings{ross_reduction_2010,
	title = {A {Reduction} of {Imitation} {Learning} and {Structured} {Prediction} to {No}-{Regret} {Online} {Learning}},
	url = {https://www.semanticscholar.org/paper/A-Reduction-of-Imitation-Learning-and-Structured-to-Ross-Gordon/79ab3c49903ec8cb339437ccf5cf998607fc313e},
	urldate = {2026-04-24},
	author = {Ross, S. and Gordon, Geoffrey J. and Bagnell, J.},
	month = nov,
	year = {2010},
}

@article{haarnoja_learning_2024,
	title = {Learning {Agile} {Soccer} {Skills} for a {Bipedal} {Robot} with {Deep} {Reinforcement} {Learning}},
	volume = {9},
	issn = {2470-9476},
	url = {http://arxiv.org/abs/2304.13653},
	doi = {10.1126/scirobotics.adi8022},
	number = {89},
	urldate = {2026-04-23},
	journal = {Science Robotics},
	author = {Haarnoja, Tuomas and Moran, Ben and Lever, Guy and Huang, Sandy H. and Tirumala, Dhruva and Humplik, Jan and Wulfmeier, Markus and Tunyasuvunakool, Saran and Siegel, Noah Y. and Hafner, Roland and Bloesch, Michael and Hartikainen, Kristian and Byravan, Arunkumar and Hasenclever, Leonard and Tassa, Yuval and Sadeghi, Fereshteh and Batchelor, Nathan and Casarini, Federico and Saliceti, Stefano and Game, Charles and Sreendra, Neil and Patel, Kushal and Gwira, Marlon and Huber, Andrea and Hurley, Nicole and Nori, Francesco and Hadsell, Raia and Heess, Nicolas},
	month = apr,
	year = {2024},
	note = {arXiv:2304.13653 [cs]},
	pages = {eadi8022},
}

@misc{tirumala_learning_2024,
	title = {Learning {Robot} {Soccer} from {Egocentric} {Vision} with {Deep} {Reinforcement} {Learning}},
	url = {http://arxiv.org/abs/2405.02425},
	doi = {10.48550/arXiv.2405.02425},
	urldate = {2026-04-23},
	publisher = {arXiv},
	author = {Tirumala, Dhruva and Wulfmeier, Markus and Moran, Ben and Huang, Sandy and Humplik, Jan and Lever, Guy and Haarnoja, Tuomas and Hasenclever, Leonard and Byravan, Arunkumar and Batchelor, Nathan and Sreendra, Neil and Patel, Kushal and Gwira, Marlon and Nori, Francesco and Riedmiller, Martin and Heess, Nicolas},
	month = may,
	year = {2024},
	note = {arXiv:2405.02425 [cs]},
}

@misc{wang_learning_2025,
	title = {Learning {Vision}-{Driven} {Reactive} {Soccer} {Skills} for {Humanoid} {Robots}},
	url = {http://arxiv.org/abs/2511.03996},
	doi = {10.48550/arXiv.2511.03996},
	urldate = {2026-04-23},
	publisher = {arXiv},
	author = {Wang, Yushi and Luo, Changsheng and Chen, Penghui and Liu, Jianran and Sun, Weijian and Guo, Tong and Yang, Kechang and Hu, Biao and Zhang, Yangang and Zhao, Mingguo},
	month = nov,
	year = {2025},
	note = {arXiv:2511.03996 [cs]},
}

@misc{chen_hifar_2025,
	title = {{HiFAR}: {Multi}-{Stage} {Curriculum} {Learning} for {High}-{Dynamics} {Humanoid} {Fall} {Recovery}},
	shorttitle = {{HiFAR}},
	url = {http://arxiv.org/abs/2502.20061},
	doi = {10.48550/arXiv.2502.20061},
	urldate = {2026-04-23},
	publisher = {arXiv},
	author = {Chen, Penghui and Wang, Yushi and Luo, Changsheng and Cai, Wenhan and Zhao, Mingguo},
	month = feb,
	year = {2025},
	note = {arXiv:2502.20061 [cs]},
}

@misc{ji_hierarchical_2022,
	title = {Hierarchical {Reinforcement} {Learning} for {Precise} {Soccer} {Shooting} {Skills} using a {Quadrupedal} {Robot}},
	url = {http://arxiv.org/abs/2208.01160},
	doi = {10.48550/arXiv.2208.01160},
	urldate = {2026-04-23},
	publisher = {arXiv},
	author = {Ji, Yandong and Li, Zhongyu and Sun, Yinan and Peng, Xue Bin and Levine, Sergey and Berseth, Glen and Sreenath, Koushil},
	month = aug,
	year = {2022},
	note = {arXiv:2208.01160 [cs]},
}

@incollection{bianchi_ball_2015,
	address = {Cham},
	title = {Ball {Dribbling} for {Humanoid} {Biped} {Robots}: {A} {Reinforcement} {Learning} and {Fuzzy} {Control} {Approach}},
	volume = {8992},
	isbn = {978-3-319-18614-6 978-3-319-18615-3},
	shorttitle = {Ball {Dribbling} for {Humanoid} {Biped} {Robots}},
	url = {http://link.springer.com/10.1007/978-3-319-18615-3_45},
	doi = {10.1007/978-3-319-18615-3_45},
	language = {en},
	urldate = {2026-04-23},
	booktitle = {{RoboCup} 2014: {Robot} {World} {Cup} {XVIII}},
	publisher = {Springer International Publishing},
	author = {Leottau, Leonardo and Celemin, Carlos and Ruiz-del-Solar, Javier},
	editor = {Bianchi, Reinaldo A. C. and Akin, H. Levent and Ramamoorthy, Subramanian and Sugiura, Komei},
	year = {2015},
	note = {Series Title: Lecture Notes in Computer Science},
	pages = {549--561},
}

@misc{hoeller_anymal_2023,
	title = {{ANYmal} {Parkour}: {Learning} {Agile} {Navigation} for {Quadrupedal} {Robots}},
	shorttitle = {{ANYmal} {Parkour}},
	url = {https://arxiv.org/abs/2306.14874v1},
	language = {en},
	urldate = {2026-04-23},
	journal = {arXiv.org},
	author = {Hoeller, David and Rudin, Nikita and Sako, Dhionis and Hutter, Marco},
	month = jun,
	year = {2023},
}

@misc{li_reinforcement_2024,
	title = {Reinforcement {Learning} for {Versatile}, {Dynamic}, and {Robust} {Bipedal} {Locomotion} {Control}},
	url = {http://arxiv.org/abs/2401.16889},
	doi = {10.48550/arXiv.2401.16889},
	urldate = {2026-04-23},
	publisher = {arXiv},
	author = {Li, Zhongyu and Peng, Xue Bin and Abbeel, Pieter and Levine, Sergey and Berseth, Glen and Sreenath, Koushil},
	month = aug,
	year = {2024},
	note = {arXiv:2401.16889 [cs]},
}

@misc{rudin_parkour_2025,
	title = {Parkour in the {Wild}: {Learning} a {General} and {Extensible} {Agile} {Locomotion} {Policy} {Using} {Multi}-expert {Distillation} and {RL} {Fine}-tuning},
	shorttitle = {Parkour in the {Wild}},
	url = {http://arxiv.org/abs/2505.11164},
	doi = {10.48550/arXiv.2505.11164},
	urldate = {2026-04-23},
	publisher = {arXiv},
	author = {Rudin, Nikita and He, Junzhe and Aurand, Joshua and Hutter, Marco},
	month = may,
	year = {2025},
	note = {arXiv:2505.11164 [cs]},
}

@misc{li_learning_2023,
	title = {Learning {Agility} and {Adaptive} {Legged} {Locomotion} via {Curricular} {Hindsight} {Reinforcement} {Learning}},
	url = {http://arxiv.org/abs/2310.15583},
	doi = {10.48550/arXiv.2310.15583},
	urldate = {2026-04-23},
	publisher = {arXiv},
	author = {Li, Sicen and Pang, Yiming and Bai, Panju and Liu, Zhaojin and Li, Jiawei and Hu, Shihao and Wang, Liquan and Wang, Gang},
	month = oct,
	year = {2023},
	note = {arXiv:2310.15583 [cs]},
}

@misc{margolis_rapid_2022,
	title = {Rapid {Locomotion} via {Reinforcement} {Learning}},
	url = {http://arxiv.org/abs/2205.02824},
	doi = {10.48550/arXiv.2205.02824},
	urldate = {2026-04-23},
	publisher = {arXiv},
	author = {Margolis, Gabriel B. and Yang, Ge and Paigwar, Kartik and Chen, Tao and Agrawal, Pulkit},
	month = may,
	year = {2022},
	note = {arXiv:2205.02824 [cs]},
}

@article{wang_cts_2024,
	title = {{CTS}: {Concurrent} {Teacher}-{Student} {Reinforcement} {Learning} for {Legged} {Locomotion}},
	volume = {9},
	issn = {2377-3766},
	shorttitle = {{CTS}},
	url = {https://ieeexplore.ieee.org/abstract/document/10670293},
	doi = {10.1109/LRA.2024.3457379},
	number = {11},
	urldate = {2026-04-22},
	journal = {IEEE Robotics and Automation Letters},
	author = {Wang, Hongxi and Luo, Haoxiang and Zhang, Wei and Chen, Hua},
	month = nov,
	year = {2024},
	pages = {9191--9198},
}

@article{hu_humanoid_2023,
	title = {Humanoid {Soccer} {Robot} {Target} {Detection} and {Localization}},
	url = {https://dl.acm.org/doi/10.1145/3632971.3632992},
	doi = {10.1145/3632971.3632992},
	language = {en},
	urldate = {2026-04-22},
	journal = {Proceedings of the 2023 International Joint Conference on Robotics and Artificial Intelligence},
	author = {Hu, Junfei and Yan, Bixi and Wang, Jun and Sun, Peng},
	month = jul,
	year = {2023},
	note = {Conference Name: JCRAI 2023: 2023 International Joint Conference on Robotics and Artificial Intelligence
ISBN: 9798400707704
Place: Shanghai China
Publisher: ACM},
	pages = {65--69},
}

@inproceedings{chen_vmts_2025,
	title = {{VMTS}: {Vision}-{Assisted} {Teacher}-{Student} {Reinforcement} {Learning} for {Multi}-{Terrain} {Locomotion} in {Bipedal} {Robots}},
	issn = {2153-0866},
	shorttitle = {{VMTS}},
	url = {https://ieeexplore.ieee.org/document/11245927/},
	doi = {10.1109/IROS60139.2025.11245927},
	urldate = {2026-04-21},
	booktitle = {2025 {IEEE}/{RSJ} {International} {Conference} on {Intelligent} {Robots} and {Systems} ({IROS})},
	author = {Chen, Fu and Wan, Rui and Liu, Peidong and Zheng, Nanxing and Wang, Bingyi and Zhou, Bo},
	month = oct,
	year = {2025},
	note = {ISSN: 2153-0866},
	pages = {4759--4766},
}

\end{document}